\documentclass[letterpaper,10pt]{article}

\usepackage{multirow} 
\usepackage{booktabs} 
\usepackage[percent]{overpic} 
\usepackage{opticameet3} 

\newcommand\authormark[1]{\textsuperscript{#1}}

\usepackage{amsmath,amssymb}
\usepackage[colorlinks=true,bookmarks=false,citecolor=blue,urlcolor=blue]{hyperref}

\usepackage[nolist]{acronym}
\usepackage{orcidlink}
\usepackage[caption=false,font=normalsize]{subfig} 

\usepackage{todonotes}
\presetkeys%
    {todonotes}%
    {inline}{}

\begin{document}

\title{Policy-driven Conformal Prediction for Trustworthy QoT Estimation}

\author{
Kiarash Rezaei \orcidlink{0009-0003-6166-2614},\authormark{1, *}
Omran Ayoub \orcidlink{0000-0002-3884-3594},\authormark{2}
Paolo Monti \orcidlink{0000-0002-5636-9910},\authormark{1}
and Carlos Natalino \orcidlink{0000-0001-7501-5547}\authormark{1}
}

\address{
\authormark{1}Department of Electrical Engineering, Chalmers University of Technology, 412 96 Gothenburg, Sweden\\
\authormark{2}University of Applied Sciences and Arts of Southern Switzerland, 6928 Lugano, Switzerland
}

\email{\authormark{*}kiarashr@chalmers.se}

\vspace{-3mm}
\begin{abstract}
We propose Conformal QoT, a policy-driven framework that combines statistically guaranteed QoT estimation with operational decision policies, enabling reliable lightpath-feasibility predictions under domain shift and improving accuracy from 92\% to 99.6\% on open datasets.
\end{abstract}

\acrodef{ZS}{zero-shot}
\acrodef{TL}{transfer learning}
\acrodef{TD}{target-domain}
\acrodef{RA}{risk-averse}
\acrodef{EO}{efficiency-oriented}
\acrodef{GC}{global conformal}
\acrodef{MU}{model uncertainty}
\acrodef{CP}{conformal prediction}
\acrodef{BV}{band validity}
\acrodef{LP}{lightpath}
\acrodef{OSNR}{optical signal-to-noise ratio}
\acrodef{ZSM}{zero-touch network management}
\acrodef{LAM}{large AI model}
\acrodef{SLA}{service level agreement}
\acrodef{QES}{quality efficiency score}
\acrodef{CQI}{Composite Quality Index}
\acrodef{CoT}{Chain-of-Thought}
\acrodef{RL}{reinforcement learning}
\acrodef{AI}{artificial intelligence}
\acrodef{AI/ML}{artificial intelligence \& machine learning}
\acrodef{ANN}{artificial neural network}
\acrodef{ASE}{amplified spontaneous emissions}
\acrodef{CUT}{channel under test}
\acrodef{DRL}{deep reinforcement learning}
\acrodef{EGN}{enhanced Gaussian noise}
\acrodef{EON}{elastic optical network}
\acrodef{GT}{ground truth}
\acrodef{GSNR}{generalized signal-to-noise ratio}
\acrodef{KDE}{kernel density estimation}
\acrodef{LLM}{large language model}
\acrodef{LP}{lightpath}
\acrodef{MAE}{mean absolute error}
\acrodef{MF}{modulation format}
\acrodef{ML}{machine learning}
\acrodef{NLI}{non-linear impairments}
\acrodef{PDF}{probability density function}
\acrodef{QoT}{quality of transmission}
\acrodef{RMSA}{routing, modulation and spectrum assignment}
\acrodef{RMSE}{root mean squared error}
\acrodef{XGB}{XGBoost}
\acrodef{IBN}{intent-based networking}
\acrodef{BER}{bit error rate}
\acrodef{XAI}{explainable artificial intelligence}
\acrodef{ADON}{autonomous driving optical network}
\acrodef{SHAP}{SHapley Additive exPlanations}

\section{Introduction}
\vspace{-2mm}

Trustworthy \ac{QoT} estimation is essential for informed optical network operations, particularly as networks grow more complex through disaggregation, spatial multiplexing (multi-core, multi-band), and tighter design margins.
In this evolving landscape, trustworthiness implies not only accuracy but also consistent and reliable predictions across domains and under varying network conditions, often with scarce available data, as improved estimation directly translates to reduced costs and extended network lifetime.
One of the most important aspects of trustworthy \ac{AI/ML} is \emph{uncertainty quantification}, which enables assessing the confidence of its predictions and identify cases where decisions should be treated with caution.
A model should also adapt to changing conditions without requiring extensive data.
Thus, a model that indicates when and where its predictions may be unreliable is inherently more trustworthy. 
Yet, most existing \ac{ML}-based models do not 
effectively quantify 
uncertainty, and this lack of quantification undermines the trust and adoption of data-driven optical network design, planning, management, and control \cite{Musumeci:25}.

Early studies framed the \ac{ML}-based \ac{QoT} estimation task as binary feasibility classification \cite{7830278, Rottondi:18}, 
later extending to uncertainty-aware approaches where neural networks were used to jointly estimate \ac{OSNR} and its uncertainty via dropout, identifying unreliable predictions \cite{Tanimura:19}. Bayesian and quantile regression further advanced probabilistic \ac{QoT} forecasting \cite{8657333}.
Calibrated regression with gradient boosting and isotonic recalibration has produced statistically consistent intervals \cite{9900791}, and probabilistic low-margin design \cite{10144068} has balanced spectrum use and risk.
However, these approaches remain model- and data-dependent, 
limiting their generalizability. 
As a result, their performance often degrades under changing network conditions or across different network domains, limiting practical use for operators who have limited data and need \ac{QoT} estimates without frequent retraining or recalibration.

The paper addresses the above challenges by proposing a two-layer, policy-driven framework, referred to as \emph{Conformal QoT}, which combines statistically guaranteed QoT estimation with its translation into operational decisions. In the first layer, \ac{CP} is applied to the \ac{ML}-based \ac{QoT} estimation, providing model-agnostic, distribution-free guarantees that remain valid under domain shift without retraining. Each QoT prediction is accompanied by a confidence band with formal guarantees, enabling trustworthy assessment of prediction reliability. The second layer introduces a decision policy 
that combines the \ac{QoT} estimation and conformal confidence bands into reliable \ac{LP} feasibility decisions, linking predictive confidence with operational action.
We assess the framework in realistic deployment scenarios where models trained on data of one transmission technology are reused with a different one. 
Results show that combining guaranteed QoT estimation with our policy-driven decision-making improves accuracy from 92\% to 99.6\% on the same dataset, enabling tighter margins that reduce infrastructure costs and extend network lifetime. 
For cross-domain deployment, a model trained on source-domain data and conformalized with only 20\% of target-domain samples achieves a 21.7\% accuracy improvement over the baseline while reducing false positives by over 99\%. 
This demonstrates that the framework enables reliable model reuse across network domains with minimal target data and without retraining, effectively bridging the gap between uncertainty estimation and operational decision-making in modern optical networks.
\vspace{-2mm}
\section{Conformal QoT Framework}
\begin{figure*}[!htbp]
  \centering
  \begin{overpic}[width=0.2\textwidth,trim=0 0 0 0,clip]{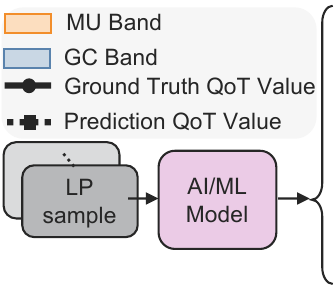}
    \put(5, 4){a)}
  \end{overpic}%
  \hspace{-0.3em}
  \begin{overpic}[width=0.23\textwidth,trim=0 0 0 0,clip]{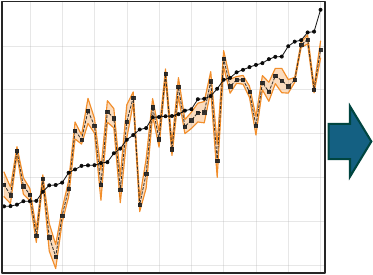}
    \put(1,4){b)}
  \end{overpic}%
  \hspace{-0.3em}
  \begin{overpic}[width=0.14\textwidth,trim=0 0 0 0,clip]{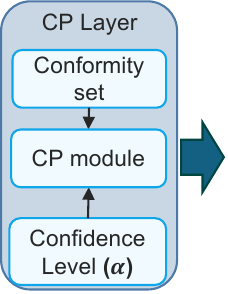}
    \put(-13,4){c)}
  \end{overpic}
  \hspace{-0.55em}
  \begin{overpic}[width=0.24\textwidth,trim=0 0 0 0,clip]{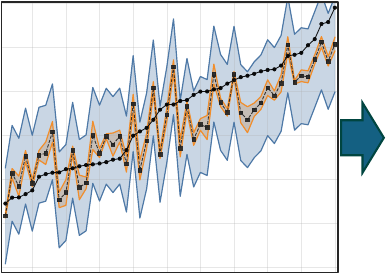}
    \put(1,4){d)}
  \end{overpic}
  \hspace{-0.55em}
  \begin{overpic}[width=0.19\textwidth,trim=0 0 0 0,clip]{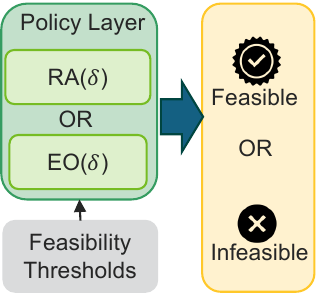}
    \put(-13,4){e)}
  \end{overpic}
  \vspace{-3mm}
  \caption{Overview of the Conformal QoT framework.}
  \label{fig:framework}
  \vspace{-2\baselineskip}
\end{figure*}

\vspace{-2mm}
\label{sec:framework}
The proposed \emph{Conformal QoT} framework (Fig. \ref{fig:framework}) comprises three parts:
\emph{(i)} a \ac{ML}-based \ac{QoT} regressor;
\emph{(ii)} a \ac{CP} that produces statistically reliable confidence bands; and 
\emph{(iii)} a policy that drives the \ac{LP} feasibility decisions.
Operations begin with a base \ac{AI/ML} regressor that predicts the \ac{QoT} metric of interest for a given \ac{LP} sample (Fig. \ref{fig:framework}(a)).
Traditionally, a \ac{MU} band (Fig. \ref{fig:framework}(b)) can be extracted from a \ac{AI/ML} regressor, quantifying how predictions vary when the model is run multiple times with different random seeds, indicating its confidence level.
However, the \ac{MU} band lacks statistical assurance, as the true \ac{QoT} value may fall outside its range. 
Moreover, the \ac{MU} band is dataset-dependent, making it unreliable under domain shift, i.e., when network conditions change.

The \emph{\ac{CP} layer} (Fig. \ref{fig:framework}(c)) overcomes these limitations by providing distribution-free statistical guarantees that remain valid even under domain shift.
It uses a small conformity set from the target domain validation data to assess deviations between predicted and true \ac{QoT} values.
This approach remains effective even with limited target-domain data, as a small conformity set still provides formal guarantees.
Using these deviations and a user-defined significance level $\alpha$ (e.g., $\alpha=0.1$ for 90\% confidence), we compute an error bound such that at least $(1-\alpha)$ of the samples lie within the bound.
This bound extends each prediction to form the \ac{GC} band (Fig.~\ref{fig:framework}(d)), ensuring that the true \ac{QoT} lies within it with probability $(1-\alpha)$, without distributional assumptions or retraining.
The \emph{policy layer} (Fig. \ref{fig:framework}(e)) translates the statistically valid \ac{GC} bands into \ac{LP} feasibility decisions through a tunable risk-tolerance parameter $\delta$, enabling operators to balance spectrum efficiency against operational reliability based on their specific requirements.
We propose two policies that leverage the \ac{GC} bands differently.
The \acf{RA} policy considers a \ac{LP} feasible only if the entire \ac{GC} band lies above the \ac{QoT} feasibility threshold, prioritizing conservative decisions that minimize incorrect acceptances (false positives).
In contrast, the \ac{EO} policy classifies a \ac{LP} as feasible when the upper bound of the band exceeds the threshold plus $\delta$, enabling more aggressive provisioning while maintaining statistical guarantees.
\vspace{-2mm}
\section{Experiments and Results}
\vspace{-2mm}
We evaluated the \emph{Conformal QoT} framework using public datasets from \cite{11018348Akbari} to demonstrate its effectiveness under domain shift.
Dataset 01 (Bu-SMF) and Dataset 07 (WC-MCF) from the CONUS topology represent source and target domains for single-core and multi-core fiber technologies, emulating scenarios where operators deploy new transmission technologies and assess whether models trained on the existing infrastructure can be reliably reused.
We adopt an \ac{XGB} model as the base regressor due to its strong performance in prior \ac{QoT} studies, training it to predict the received \ac{OSNR} of each \ac{LP} configuration.
Point predictions serve as the baseline policy, where decisions rely solely on predicted \ac{OSNR} values and feasibility thresholds without uncertainty quantification.

Three training regimes were evaluated to assess cross-domain generalization capabilities: \acf{ZS}, \acf{TL}, and \acf{TD}.
In \ac{ZS}, the model is trained on 80\% of source domain samples (single-core) and evaluated directly on the target domain (multi-core), with 20\% of target data reserved for the conformity set, simulating operators reusing existing models on new infrastructure without retraining.
For \ac{TL}, the model is pre-trained on the source domain, then fine-tuned using 20\% of target data.
In \ac{TD}, the model is trained and tested entirely within the target domain using 90\% for training and 10\% for the conformity set.
All data splits were mutually exclusive to prevent leakage. 

\begin{table}[hbt]
\centering
\vspace{-6mm}
\caption{Regression (\acf{MAE} and \acf{RMSE}) and uncertainty (\acf{MU} and \acf{GC}) metrics across training regimes.}
\label{tab:cp_bands_and_table}
{\small
\renewcommand{\arraystretch}{0.9}
\setlength{\tabcolsep}{3pt}
\begin{tabular}{lcccccccc}
\toprule
\multirow{2}{*}{Model} 
 & \multicolumn{2}{c}{Regression Metrics} 
 & \multicolumn{2}{c}{\ac{MU}} 
 & \multicolumn{2}{c}{\ac{GC}} \\
\cmidrule(lr){2-3} \cmidrule(lr){4-5} \cmidrule(lr){6-7}
 & MAE [dB] & RMSE [dB] 
 & Conf. [\%] & Width [dB] 
 & Conf. [\%] & Width [dB] \\
\midrule
\Acf{ZS} & 1.97 & 1.98 & 0  & 0.04 & 90 & 4.17 \\
\Acf{TL} & 0.03 & 0.03 & 18 & 0.01 & 90 & 0.11 \\
\Acf{TD} & 0.02 & 0.03 & 32 & 0.02 & 91 & 0.10 \\
\bottomrule
\end{tabular}
}
\vspace{-3mm}
\label{tab:cp_bands_and_table}
\end{table}

Table \ref{tab:cp_bands_and_table} shows that while \ac{XGB} achieves excellent accuracy in \ac{TL} and \ac{TD} regimes (sub-dB \ac{MAE}), the \ac{ZS} model shows significant degradation due to domain mismatch.
The \ac{MU} bands prove unreliable across all regimes, achieving only 0\%, 18\%, and 32\% confidence despite narrow widths that misleadingly suggest certainty.
This demonstrates why traditional uncertainty estimates fail under domain shift and cannot be trusted for operational decisions.
In contrast, conformalized \ac{GC} bands consistently deliver 90–91\% confidence across all regimes.
For \ac{TL} and \ac{TD}, these guarantees come with minimal band widths, enabling tight margins that translate into reliable \ac{LP} decisions.
Even in the challenging \ac{ZS} scenario, the framework provides valid guarantees with wider bands (4.17 dB), reflecting inherent cross-domain uncertainty.

\begin{figure}
    \centering
\includegraphics[width=0.75\textwidth]
{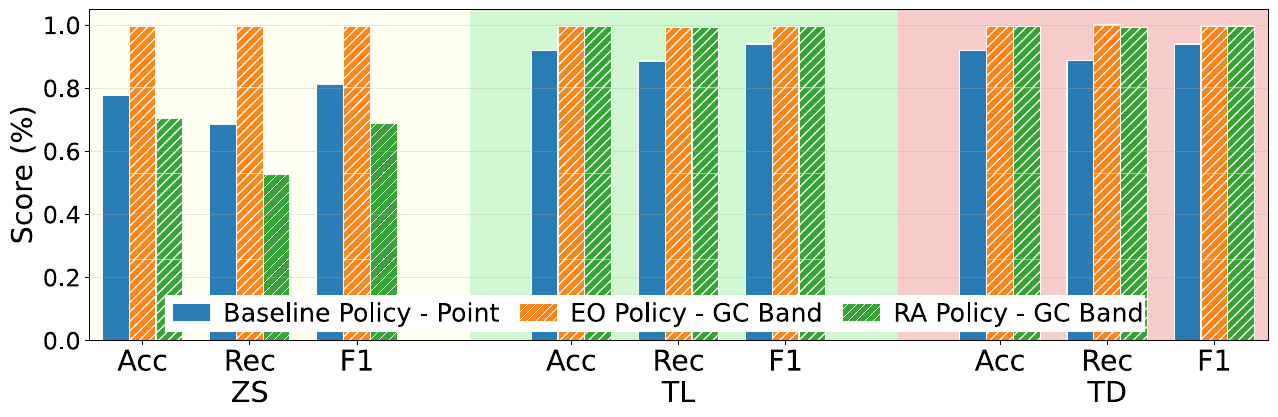}
\vspace{-0.5cm} %
\caption{Accuracy (Acc), recall (Re), and F1-score (F1) across training regimes.}
\label{fig:performance_barplot}
\vspace{-0.6cm}
\end{figure}

To assess the reliability of \ac{LP} feasibility decisions, each test \ac{LP} was evaluated with its assigned modulation format and associated \ac{OSNR} threshold.
Table \ref{tab:confusion_matrix} presents detailed classification metrics, while Fig. \ref{fig:performance_barplot} summarizes overall performance.
True positive rate (TPR) and true negative rate (TNR) indicate correctly feasible and correctly infeasible \acp{LP}, while false positive rate (FPR) and false negative rate (FNR) capture incorrect decisions leading to either failed lightpath establishments (operational disruptions) or missed provisioning opportunities (lower spectral efficiency or revenue loss).

In Table \ref{tab:confusion_matrix}, the operational impact is most evident in \ac{ZS}, where baseline predictions yield high FPR, i.e., nearly one in four feasible lightpaths would fail.
The \ac{EO} policy reduces this failure rate by over 99\% (to 0.3\%) while more than doubling TPR (from 29.4\% to 62.2\%).
The \ac{RA} policy eliminates false acceptances (0\% FPR) with moderate conservatism (FNR $\approx$ 29.7\%), offering maximum operational safety.
This demonstrates that models trained on existing infrastructure can be reliably deployed on new technologies using only a small conformity set, eliminating the need for retraining and enabling rapid technology adoption.
For \ac{TL} and \ac{TD} models, both policies maintain high TPR (around 62\%) with near-zero FNR while reducing FPR from 8\% to below 0.5\%.
This transforms nominally accurate models (92\% baseline accuracy) into highly reliable systems (99.6\% accuracy), enabling tighter margins that extend network lifetime and reduce infrastructure costs.
The improvement from 77.7\% to 99.4\% accuracy in cross-domain \ac{ZS} is particularly significant, confirming the framework enables model reuse across network domains with limited target data and without retraining.
Fig. \ref{fig:performance_barplot} shows that conformalized policies consistently improve F1-score, recall, and accuracy across all regimes.
Notably, the conformalized \ac{ZS} model with \ac{EO} policy achieves performance comparable to \ac{TL} and \ac{TD} models, demonstrating that statistical guarantees through conformal prediction bridge the domain gap and enable reliable zero-shot deployment, critical as optical networks evolve toward disaggregated, multi-band, multi-core architectures where training data is scarce.
Operators can adjust $(1-\alpha)$ and $\delta$ to balance reliability against efficiency based on their operational requirements.

\begin{table}[!t]
    \centering
    {\small
\renewcommand{\arraystretch}{0.9}
\setlength{\tabcolsep}{3pt}
\caption{Performance results and risk-tolerance parameters~$\delta$.}
\label{tab:confusion_matrix}
\vspace{-2mm}
\begin{tabular}{l|cccc|cccc|cccc|cc}
\toprule
\multirow{2}{*}{Model} 
 & \multicolumn{4}{c|}{Point + Baseline} 
 & \multicolumn{4}{c|}{\ac{GC} + \ac{EO}($\delta_{\text{EO}}$)} 
 & \multicolumn{4}{c|}{\ac{GC} + \ac{RA}($\delta_{\text{RA}}$)} 
 & \multirow{2}{*}{$\delta_{\text{EO}}$} 
 & \multirow{2}{*}{$\delta_{\text{RA}}$} \\ 
\cmidrule(lr){2-13}
 & TPR & TNR & FPR & FNR & TPR & TNR & FPR & FNR & TPR & TNR & FPR & FNR \\
\midrule
\Acf{ZS} & 29.4 & 48.3 & 22.3 & 0.0 & 62.2 & 37.2 & 0.3 & 0.3 & 32.9 & 37.5 & 0.0 & 29.7 & 0.1 & 0.0 \\
\Acf{TL} & 29.4 & 62.5 & 8.0 & 0.0 & 62.1 & 37.5 & 0.0 & 0.4 & 62.0 & 37.5 & 0.0 & 0.5 & 0.1 & 0.0 \\
\Acf{TD} & 29.4 & 62.6 & 8.0 & 0.0 & 62.5 & 37.1 & 0.4 & 0.0 & 62.1 & 37.5 & 0.0 & 0.4 & 0.0 & 0.0 \\
\bottomrule
\end{tabular}
}
\vspace{-0.2cm}
\end{table}
%
\vspace{-2mm}
\section{Conclusions}
\vspace{-2mm}
We combine \ac{CP} with policy-driven decision logic to provide statistically guaranteed confidence bands that remain valid under domain shift without retraining. It reduces \ac{LP} decision errors and enables reliable cross-domain model reuse. By delivering trustworthy \ac{QoT} estimation with minimal target-domain data, the framework supports rapid and reliable technology deployment in modern optical networks.

\vspace{2pt}
\noindent{\textbf{\small Acknowledgments:}} \footnotesize 
This work has been supported by the Horizon Europe ECO-eNET project with funding from the SNS JU under grant agreement No. 101139133.
\vspace{-2mm}
\bibliographystyle{opticaconf} 
\bibliography{references}

@ARTICLE{11018348Akbari,
  author={Akbari, Hassan and Shariati, Behnam and Moreno Morrone, Juan L. and Safari, Pooyan and Fischer, Johannes K. and Freund, Ronald},
  journal={Journal of Optical Communications and Networking}, 
  title={Datasets for QoT estimation in SDM networks}, 
  year={2025},
  volume={17},
  number={6},
  pages={514-525},
  doi={10.1364/JOCN.558452}}

@article{Rottondi:18,
author = {Cristina Rottondi and Luca Barletta and Alessandro Giusti and Massimo Tornatore},
journal = {J. Opt. Commun. Netw.},
number = {2},
pages = {A286--A297},
publisher = {Optica Publishing Group},
title = {Machine-Learning Method for Quality of Transmission Prediction of Unestablished Lightpaths},
volume = {10},
month = {Feb},
year = {2018},
url = {https://opg.optica.org/jocn/abstract.cfm?URI=jocn-10-2-A286},
doi = {10.1364/JOCN.10.00A286},
}

@INPROCEEDINGS{9900791,
  author={Di Cicco, Nicola and Ibrahimi, Mëmëdhe and Rottondi, Cristina and Tornatore, Massimo},
  booktitle={Proc. of BalkanCom}, 
  title={Calibrated Probabilistic QoT Regression for Unestablished Lightpaths in Optical Networks}, 
  year={2022},
  volume={},
  number={},
  pages={21-25},
  doi={10.1109/BalkanCom55633.2022.9900791}}

@ARTICLE{10144068,
  author={Karandin, Oleg and Ferrari, Alessio and Musumeci, Francesco and Pointurier, Yvan and Tornatore, Massimo},
  journal={Journal of Optical Communications and Networking}, 
  title={Probabilistic low-margin optical-network design with multiple physical-layer parameter uncertainties}, 
  year={2023},
  volume={15},
  number={7},
  pages={C129-C137},
  doi={10.1364/JOCN.482734}}

@article{Musumeci:25,
author = {Francesco Musumeci and Massimo Tornatore},
journal = {J. Opt. Commun. Netw.},
number = {8},
pages = {C144--C155},
publisher = {Optica Publishing Group},
title = {Failure management in optical networks with ML: a tutorial on applications, challenges, and pitfalls \[Invited\]},
volume = {17},
month = {Aug},
year = {2025},
url = {https://opg.optica.org/jocn/abstract.cfm?URI=jocn-17-8-C144},
doi = {10.1364/JOCN.551910},
}

@ARTICLE{8657333,
  author={Meng, Fanchao and Mavromatis, Alex and Bi, Yu and Wang, Rui and Yan, Shuangyi and Nejabati, Reza and Simeonidou, Dimitra},
  journal={Journal of Optical Communications and Networking}, 
  title={Self-learning monitoring on-demand strategy for optical networks}, 
  year={2019},
  volume={11},
  number={2},
  pages={A144-A154},
  doi={10.1364/JOCN.11.00A144}}

@ARTICLE{7830278,
  author={Panayiotou, T. and Chatzis, S. P. and Ellinas, G.},
  journal={Journal of Optical Communications and Networking}, 
  title={Performance analysis of a data-driven quality-of-transmission decision approach on a dynamic multicast- capable metro optical network}, 
  year={2017},
  volume={9},
  number={1},
  pages={98-108},
  doi={10.1364/JOCN.9.000098}}

@ARTICLE{Tanimura:19,
  author={Tanimura, Takahito and Hoshida, Takeshi and Kato, Tomoyuki and Watanabe, Shigeki},
  journal={Journal of Lightwave Technology}, 
  title={{OSNR} Estimation Providing Self-Confidence Level as Auxiliary Output From Neural Networks}, 
  year={2019},
  volume={37},
  number={7},
  pages={1717-1723},
  doi={10.1109/JLT.2019.2895730},
}

\acrodef{ZS}{zero-shot}
\acrodef{TL}{transfer learning}
\acrodef{TD}{target-domain}
\acrodef{RA}{risk-averse}
\acrodef{EO}{efficiency-oriented}
\acrodef{GC}{global conformal}
\acrodef{MU}{model uncertainty}
\acrodef{CP}{conformal prediction}
\acrodef{BV}{band validity}
\acrodef{LP}{lightpath}
\acrodef{OSNR}{optical signal-to-noise ratio}
\acrodef{ZSM}{zero-touch network management}
\acrodef{LAM}{large AI model}
\acrodef{SLA}{service level agreement}
\acrodef{QES}{quality efficiency score}
\acrodef{CQI}{Composite Quality Index}
\acrodef{CoT}{Chain-of-Thought}
\acrodef{RL}{reinforcement learning}
\acrodef{AI}{artificial intelligence}
\acrodef{AI/ML}{artificial intelligence \& machine learning}
\acrodef{ANN}{artificial neural network}
\acrodef{ASE}{amplified spontaneous emissions}
\acrodef{CUT}{channel under test}
\acrodef{DRL}{deep reinforcement learning}
\acrodef{EGN}{enhanced Gaussian noise}
\acrodef{EON}{elastic optical network}
\acrodef{GT}{ground truth}
\acrodef{GSNR}{generalized signal-to-noise ratio}
\acrodef{KDE}{kernel density estimation}
\acrodef{LLM}{large language model}
\acrodef{LP}{lightpath}
\acrodef{MAE}{mean absolute error}
\acrodef{MF}{modulation format}
\acrodef{ML}{machine learning}
\acrodef{NLI}{non-linear impairments}
\acrodef{PDF}{probability density function}
\acrodef{QoT}{quality of transmission}
\acrodef{RMSA}{routing, modulation and spectrum assignment}
\acrodef{RMSE}{root mean squared error}
\acrodef{XGB}{XGBoost}
\acrodef{IBN}{intent-based networking}
\acrodef{BER}{bit error rate}
\acrodef{XAI}{explainable artificial intelligence}
\acrodef{ADON}{autonomous driving optical network}
\acrodef{SHAP}{SHapley Additive exPlanations}
\end{document}